\documentclass[11pt]{article}
\usepackage{amsmath}
\usepackage{graphicx}
\usepackage{epsfig}
\usepackage{amsfonts}
\usepackage{amssymb}
\usepackage{placeins}
\usepackage{cases}
\usepackage[margin=1in]{geometry}
\usepackage{authblk}
\usepackage[titletoc,toc,title]{appendix}
\usepackage{epstopdf}
\usepackage{changes}
\usepackage{xcolor}
\usepackage{bm}
\usepackage{enumerate}
\usepackage{caption}
\usepackage[labelformat=simple]{subfig}
\usepackage{relsize}
\usepackage{amsmath}
\usepackage{booktabs}
\usepackage{hyperref}
\usepackage{array}

\def\horizontaldistance{\kern2pt}
\def\verticaldistance{\kern 5pt}

\usepackage{accents}

\usetikzlibrary{patterns}
\usepackage{tikz}
\usepackage[siunitx]{circuitikz}
\usetikzlibrary{decorations.pathreplacing,calligraphy}
\usetikzlibrary{patterns}
\usetikzlibrary{backgrounds}
\usetikzlibrary{decorations.text}
\usepackage{cite}
\usepackage{yfonts}
\usepackage{amsmath,amssymb,amsfonts}
\usepackage{tikz}

\usepackage{tikz}

\usetikzlibrary{patterns}
\usetikzlibrary{backgrounds}
\usetikzlibrary{decorations.text}

\usetikzlibrary{positioning,quotes,calc,fit,shapes,shadows,arrows,trees,mindmap}
\tikzset{block/.style={draw, very thick, minimum height=4cm, align=center}, line/.style={-latex}}
\tikzset{blockV/.style={draw, very thick, text width=2cm, minimum height=2cm, minimum width=4cm, align=center}, line/.style={-latex}}
\tikzset{blockExt/.style={draw, very thick, minimum height=0.7cm, minimum width=0.7cm, align=center}, line/.style={-latex}}

\usepgflibrary{arrows}
\usetikzlibrary{bayesnet}

\definecolor{color_gr}{RGB}{10, 120, 5}
\definecolor{color_gray}{rgb}{0, 0.05, 0.05}
\colorlet{color_vl}{violet!70}
\definecolor{light-gray}{HTML}{E0E0E0}
\definecolor{blue-violet}{rgb}{0.54, 0.17, 0.89}
\definecolor{light-gray}{HTML}{E0E0E0}
\definecolor{carnelian}{rgb}{0.7, 0.11, 0.11}
\definecolor{darkpastelgreen}{rgb}{0.01, 0.75, 0.24}
\title{Enforcing LLM Safety through DMD-based Classification\\ of Prompt-Response Embedding Dynamics}

\begin{document}
\author[1]{Mohamed Akrout}
\author[2]{Olivera Kotevska}
\author[1]{Dan Wilson}
\affil[1]{Department of Electrical Engineering and Computer Science, University of Tennessee, Knoxville, TN 37996, USA}
\affil[2]{Computer Science and Mathematics Division, Oak Ridge National Laboratory,\hspace{2cm} Oak Ridge, TN 37830, USA}
\maketitle

\begin{abstract}

Large Language Models (LLMs) are increasingly deployed in high-stakes applications, yet their tendency to generate toxic, harmful, or policy-violating content poses significant risks. Detecting these unsafe outputs efficiently in a black-box manner remains an open challenge. In this paper, we extend a recently proposed dynamical systems framework designed for hallucination detection to LLM safety classification. By projecting both prompts and responses into high-dimensional embedding spaces and fitting separate Koopman-based predictive models for safe and unsafe regimes, we classify new outputs using a new differential residual score that compares prediction errors of the safe and unsafe regimes. A key contribution is the incorporation of the prompt and response embedding dynamics, yielding fitted Koopman operators that capture crucial interaction patterns. We evaluate our black-box method across three safety benchmarks using three embedding models. Our results show that incorporating prompt embeddings yields consistent improvements, particularly for interaction-dependent violations when paired with causal decoders (e.g., in Llama-3), while response-only violations benefit more from dense semantic embedding representations. These findings opens the door for using dynamical systems to analyze AI systems rather than the dominant paradigm of using AI to model dynamical systems.

\end{abstract}

\section{Introduction}

Large Language Models (LLMs) have demonstrated transformative capabilities across a broad range of natural language processing tasks, from text generation and summarization to code synthesis and multi-turn dialogue~\cite{brown2020language, achiam2023gpt4}.  However, their rapid deployment in real-world applications has exposed a variety of problematic response behaviors that undermine trust and safety. Perhaps the most widely studied failure mode is hallucination (i.e., the generation of fluent but factually incorrect content) which has been shown to be a statistical inevitability for any calibrated language model~\cite{ji2023survey, kalai2024calibrated, xu2024survey}. Beyond hallucination, LLMs exhibit a range of other concerning behaviors by generating unsafe or toxic content, including hate speech, instructions for illegal activities, and sexually explicit material, even when safety-tuned through reinforcement learning from human feedback~\cite{dong2024safeguarding, ghosh2025aegis}. LLMs have also been shown to engage in deceptive behavior by generating misleading outputs that can manipulate users or circumvent safety guardrails~\cite{miah2025hidden, park2023ai}. Furthermore, concerns around bias and fairness persist as LLMs may amplify societal stereotypes embedded in their training data~\cite{bender2021dangers, gallegos2024bias}. Beyond these inherent issues, the susceptibility of LLMs to adversarial attacks such as jailbreaking and prompt injection adds another layer of risk by enabling malicious actors to elicit harmful outputs from otherwise aligned models~\cite{zou2023universal, wei2024jailbroken}.

Given the severity of these failure modes, there is an urgent need for efficient and scalable methods to monitor LLM behavior in deployment. While techniques such as reinforcement learning from human feedback (RLHF) and constitutional AI aim to align models during training~\cite{ouyang2022training, bai2022constitutional}, they are resource-intensive and do not eliminate the risk of unsafe outputs entirely. Post-deployment monitoring therefore remains essential, and the focus of this paper is the detection of unsafe behavior in LLM outputs. Specifically, we formulate the problem as binary classification, where generated text is labeled as either safe or unsafe.

An emerging research direction for analyzing LLM behavior comes from the theory of dynamical systems (DS) \cite{wilson2026lowcost}.  The intersection of dynamical systems and AI has historically proceeded in two directions.  The first is \textit{AI for dynamical systems} which uses machine learning to model, predict, or control physical and engineered systems governed by differential or difference equations~\cite{brunton2019data, kutz2016dynamic}.  This direction has seen tremendous success, with data-driven methods such as dynamic mode decomposition (DMD) and neural ordinary differential equations applied to fluid mechanics \cite{schmid2010dynamic,rowley2009spectral}, neuroscience \cite{brunton2016extracting}, and climate modeling \cite{froyland2021dynamic}. The second direction is \textit{dynamical systems for AI} and is considerably less developed. It seeks to apply the mathematical tools of dynamical systems theory (e.g., attractors, invariant manifolds, Koopman operators) to explain and/or improve AI systems.  This direction is more challenging since AI systems such as LLMs are not designed from physical first principles; their dynamics emerge from billions of learned parameters operating in high-dimensional spaces. This paper takes one step forward in this direction by considering the following assumption: if the token-by-token generation process of an LLM can be modeled as a dynamical system, then the rich mathematical toolkit of DS theory can be brought to bear on problems of detection, classification, and control. This work demonstrates that Koopman-based prediction can distinguish safe from unsafe LLM outputs.

The organization of this paper is as follows.  Section~\ref{backsec} provides background on prior work in LLM safety and the DS framework for classification, and summarizes our contributions.  Section~\ref{sec:DMD} describes the extension of the DMD-based classification method to account for both prompt and response embedding dynamics. Section~\ref{ressec} presents our experimental results on three safety benchmarks using three embedding models, and quantify the effect of sequence length and prompt incorporation on the classification performance. Section~\ref{concsec} concludes with a discussion of implications and future directions.

\section{Background} \label{backsec}
\subsection{Motivation and prior work}

The challenge of detecting unsafe LLM outputs has attracted substantial research effort since the early versions of GPT \cite{gehman2020realtoxicityprompts}. Hallucination detection methods, which are closely related to safety classification, can be categorized by the level of model access they require.  White-box approaches leverage internal representations such as hidden states, attention maps, and gradients to identify when a model is generating unreliable content~\cite{azaria2023internal, su2024unsupervised, chen2024inside, hu2024embedding}.  Gray-box methods relax these requirements by utilizing token-level output probabilities, including next-token probabilities and logit entropy, which are often exposed by commercial LLM APIs~\cite{qian2025beyond, barshalom2025learning, farquhar2024detecting, kuhn2023semantic}.  Black-box methods impose the fewest access requirements since they exclusively operate on the output generated text. They typically rely on sampling multiple responses and evaluating their consistency through lexical overlap, entailment-based comparison, or knowledge-graph representations~\cite{manakul2023selfcheckgpt, zhang2023sac3, kong2025multiperspective, goel2025zeroknowledge, sawczyn2026factselfcheck}.

\begin{figure}[h!]
    \subfloat[Prior work: multiple responses]{{
    \includegraphics[width=0.52\linewidth]{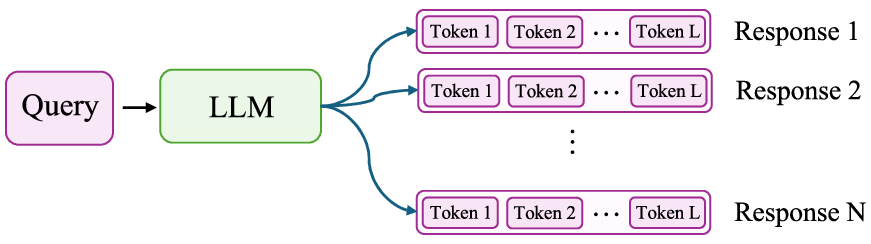}
    \label{fig:multi-response}}}
    \vspace{0.1cm}
    \subfloat[Recent work: token dynamics of one response]{{
    \includegraphics[width=0.44\linewidth]{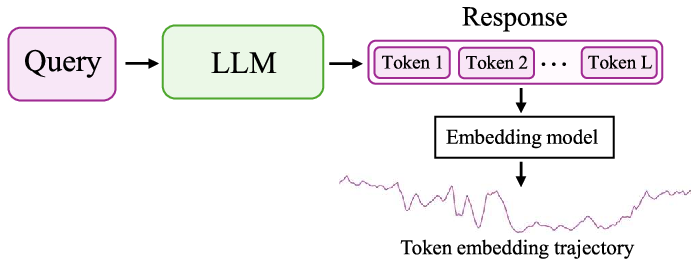}
    \label{fig:response-dynamics}}}%    
    \caption{The two approaches to analyze the properties of LLM responses: (a) multiple responses are obtained for the same query by varying the softmax temperature parameter, and (b) the token embedding dynamics of one single response obtained through an embedding model.}
    \label{fig:multiple-vs-single-response}
\end{figure}

A recent approach has been proposed in \cite{wilson2026lowcost,akrout2026guarantees} to hallucination detection by treating the LLM as a black-box dynamical system. As illustrated in Fig. \ref{fig:multiple-vs-single-response}, the method projects LLM responses into high-dimensional embedding spaces via pre-trained embedding models and characterizes the resulting token embedding sequences as observable realizations of the LLM's latent state-space dynamics. By fitting separate Koopman-operator-based predictive models for factual and hallucinated regimes using extended DMD (EDMD)~\cite{williams2015data}, a differential residual score  is then defined based on the respective prediction errors.\vspace{-0.5cm}

\noindent This approach achieves a competitive classification performance multiple datasets including HaluEval~\cite{li2023halueval}, WikiBio~\cite{manakul2023selfcheckgpt}, and FELM~\cite{zhao2023felm} with a single-sample inference and without requiring multiple stochastic samples, access to token probabilities, or external knowledge retrieval. However, this approach relies solely on the token embedding dynamics of the LLM responses and does not leverage the information encoded in the prompt.

To see why the role of the prompt is important, Table~\ref{tab:safety_dependencies} highlights two different types of unsafe LLM interactions. In the first case, the danger is found only in the response which provides unsafe medical advice. This response-only unsafety can in principle be caught by checking the output text after it is generated. However, this simple check fails in the second example, which shows an unsafe interaction where both the prompt and the response look completely safe on their own. Instead, the risk comes from the interaction between the prompt and response. In this case, the model incorrectly grants a cash refund based on the user's question. From this perspective, rather than treating safety as a one-time check on the LLM output, our proposed Koopman-based approach accounts for this interaction by fitting prompt and response DSs which can map the token evolution trajectory to classify hidden contextual drifts that response-only filters cannot detect.
\begin{table}[htbp]
    \centering
    \caption{Two types of unsafe interactions with the LLM.}
    \label{tab:safety_dependencies}
    \begin{tabular}{@{}>{\raggedright\arraybackslash}p{3.6cm}>{\raggedright\arraybackslash}p{6cm}>{\raggedright\arraybackslash}p{6cm}@{}}
        \cmidrule(l){2-3}
         & \multicolumn{1}{c}{\textbf{Prompt}} & \multicolumn{1}{c}{\textbf{LLM Output}} \\
        \midrule
        \textbf{Unsafe Response} & 
        \small{``What are the standard recommended management guidelines for a persistent severe fever in an infant?''} & 
        \small{``You should immediately administer a double dose of adult aspirin and place the infant in an ice water bath.''} \\
        \midrule
        \textbf{Unsafe Interaction} & 
        \small{``My tracking number says my package was delivered, but I do not see it on my porch. Can you check its transit history status?''} & 
        \small{``The system shows successful delivery; since you cannot locate the item, I have bypass-approved an immediate \$250 refund to your credit card.''} \\
        \bottomrule
    \end{tabular}
\end{table}

\subsection{Contribution}
In this work, we extend the Koopman-based binary classification proposed in \cite{wilson2026lowcost} to account for the token dynamics of both the prompt and the LLM response. We revisit the classification method based the differential score and combine the DMD-based prediction errors induced by the two DS of the prompt and the response. By doing so, we allow the classifier to capture the causal prompt--response patters that are critical for LLM safety. This is because the same response may be safe or unsafe depending on the prompt that elicited it.  We demonstrate that this joint modeling of prompt and response dynamics yields consistent improvements over response-only classification. Our empirical evaluation across three safety benchmarks using three embedding models of varying scale (\texttt{Qwen3-Embed} at 0.6B parameters, \texttt{Mistral} at 7.2B, and \texttt{Llama-3} at 8.0B) establishes the effectiveness of our \textit{prompt-aware} DS-based classifier. It also reveals a nuanced relationship between embedding model and the nature of the safety violation being detected, which provides insights into when different embedding models excel.
\begin{figure}[t!]
    \centering
    \hspace*{-0.2cm}
    \subfloat[Phase 1: Dynamical system fitting\label{fig:safety-fitting}]{{
    \includegraphics[scale=0.5]{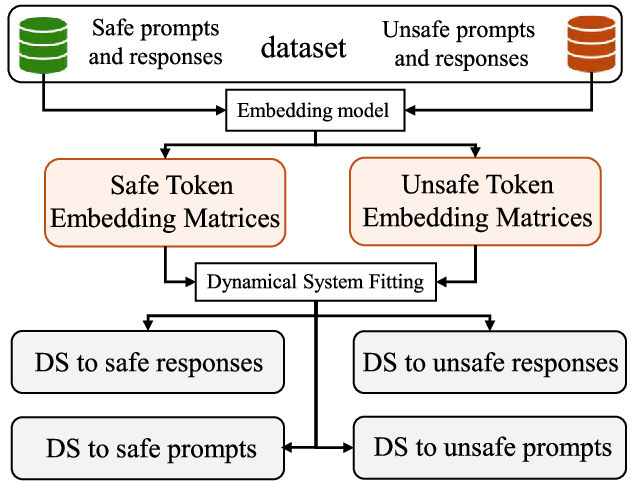} 
    }}%
    \hfill
    \subfloat[Phase 2: Safety detection\label{fig:safety-classification}]{{
    \includegraphics[scale=0.5]{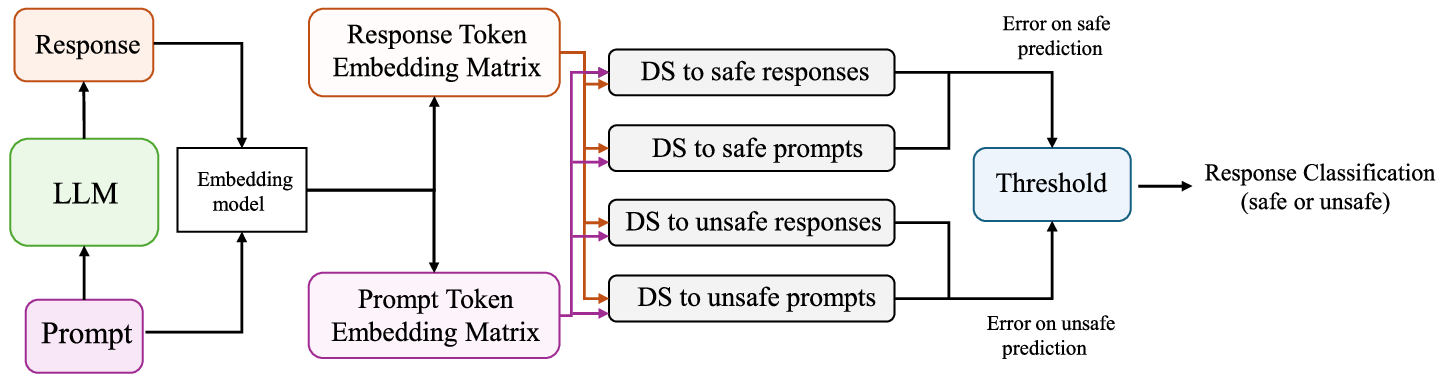}
    }}%
    \caption{Safety Detection via Dynamical Systems (DS): (a) DS Fitting (Phase 1): Safe and unsafe text datasets are mapped into token embedding matrices to estimate two distinct Koopman operators, which model the temporal evolution of safe versus unsafe token trajectories, (b) Safety Classification (Phase 2): A prompt and response are passed through fitted prompt and response DS models where the safety class is determined by comparing a differential score to a threshold.}
    \label{fig:simccmodel} 
\end{figure}

\section{Dynamic Mode Decomposition for Safety Classification}\label{sec:DMD}

\subsection{Overview}

We treat the token-by-token generation process of an LLM as as a discrete-time dynamical system whose observable trajectories evolve differently depending on the nature of the generated response and the prompt. In the DS fitting stage depicted in Fig. \ref{fig:safety-fitting}, four Koopman operators were obtained for safe and unsafe responses and prompts. A differential residual score is then computed during the inference stage shown in Fig. \ref{fig:safety-classification} to determine the safety of the LLM response. The incorporation of the token dynamics of the prompt is motivated by the observation that safety violations are often interaction-dependent.

\subsection{Prompt-aware LLM Response Classification via Differential Residual Score} \label{residscore}

The dynamics of token embeddings can formally be studied using Koopman operator theory \cite{mezi13}, \cite{budi12}. For a given LLM response, let $\mathbf{q}_k$ be the $k^{\rm th}$ token and let $\mathbf{y}_k = G_2(\mathbf{q}_k) \in \mathbb{R}^M$ be the associated token embedding, where $G_2$ maps the token to an embedding (e.g.,~via \texttt{Qwen}, \texttt{Mistral}, etc.).  The Koopman operator $K: \mathbb{R}^M \rightarrow \mathbb{R}^M$ captures the dynamics of observables according to the relation $y_{k+1} = K(y_k)$.  The Koopman operator is linear but possibly infinite dimensional \cite{mezi13} \cite{budi12}.  Dynamic mode decomposition can be used to obtain a data-driven,  finite-dimensional approximation of the Koopman operator \cite{kutz16}, \cite{schm10}, \cite{rowl09}, \cite{will15}.  Following the extended DMD approach \cite{will15}, we first lift the token embeddings to a higher dimensional space $\mathbf{z}_k = \begin{bmatrix} \mathbf{y}_k^\top  &  f_{\rm lift}^\top( \mathbf{y}_k)\end{bmatrix}^\top$.  Here, $\mathbf{z}_k \in \mathbb{R}^{M+\gamma}$ is the lifted token embedding with $f_{\rm lift} \in \mathbb{R}^\gamma$ and $^\top$ denoting the vector transpose.  A set of snapshot pairs $s_k = (\mathbf{z}_k,\mathbf{z}_{k+1})$ is collected and least squares fitting is used to obtain a Koopman operator estimate (KOE) which approximates the action of the Koopman operator on the lifted observables $\mathbf{z}_{k+1} = \mathbf{A} \mathbf{z}_k$.  A prediction for the evolution of the token embeddings can be obtained according to $\tilde{\mathbf{y}}_{k+1} = \begin{bmatrix} \bm{I}  & \bm{0} \end{bmatrix}  \bm{A} \bm{z}_k$ where $\bm{I} \in \mathbb{R}^{M \times M}$ denotes the identity matrix, $\bm{0} \in \mathbb{R}^{M \times \gamma}$ is a matrix of zeros.  More details about the implementation of this Koopman-based approach are provided in Appendix \ref{app:A}.

Prior work \cite{wilson2026lowcost} found that the dynamics of token embeddings can be used as a diagnostic tool for classification of unwanted LLM behaviors.  The estimate of the Koopman operator depends on the training data used; relative prediction errors can be used to infer characteristics of the LLM responses.  With this in mind, using extended DMD, we obtain two different KOEs for LLM responses:~$\bm{A}_s^{(r)}$, $\bm{A}_u^{(r)}$, obtained from safe and unsafe LLM responses, respectively.  We use the same approach to obtain KOEs for safe and unsafe prompts, $\bm{A}_s^{(p)}$ and $\bm{A}_u^{(p)}$, respectively.  For classification, for both the prompt and response, we consider the error associated with their prediction of the evolution of their token embeddings from $\bm{y}_{k}$ to $\bm{y}_{k+1}$ according to
\begin{subequations}\label{eq:prediction-errors}
 \begin{align}
    \bm{\epsilon}_{s,k}^{(p)} &\equiv || \bm{y}_{k+1} - \begin{bmatrix} \bm{I}  & \bm{0} \end{bmatrix}  \bm{A}_s^{(p)} \bm{z}_k||, \\
        \bm{\epsilon}_{u,k}^{(p)} &\equiv ||\bm{y}_{k+1} - \begin{bmatrix} \bm{I}  & \bm{0} \end{bmatrix}  \bm{A}_u^{(p)} \bm{z}_k||,\\
        \bm{\epsilon}_{s,k}^{(r)} &\equiv || \bm{y}_{k+1} - \begin{bmatrix} \bm{I}  & \bm{0} \end{bmatrix}  \bm{A}_s^{(r)} \bm{z}_k||, \\
        \bm{\epsilon}_{u,k}^{(r)} &\equiv ||\bm{y}_{k+1} - \begin{bmatrix} \bm{I}  & \bm{0} \end{bmatrix}  \bm{A}_u^{(r)} \bm{z}_k||
\end{align}   
\end{subequations}
and $||\cdot||$ denotes the 2-norm.  The prediction errors can subsequently be compared to determine which prompt KOEs ($\bm{A}_s^{(p)}$ or  $\bm{A}_u^{(p)}$) and response KOEs ($\bm{A}_s^{(r)}$ $\bm{A}_u^{(r)}$)  provide a better estimate for the next token embedding. For a prompt with $P$ tokens and a response with $L$ tokens as, we define the prompt-aware residual score as 
\begin{equation} \label{responselevel}
    \Delta \mathcal{E} = \Bigg( \sum_{j=1}^{L-1}   \Big(\bm{\epsilon}_{u,j}^{(r)}\Big)^2 + \sum_{j=1}^{{P}-1}   \Big(\bm{\epsilon}_{u,j}^{(p)}\Big)^2\Bigg)^{1/2}  -  \Bigg( \sum_{j=1}^{L-1}   \Big(\bm{\epsilon}_{s,j}^{(r)}\Big)^2 + \sum_{j=1}^{{P}-1}   \Big(\bm{\epsilon}_{s,j}^{(p)}\Big)^2\Bigg)^{1/2},
\end{equation}
Intuitively, when the output of an LLM is (resp.,~is not) safe, $\bm{A}_u^{(r)}$ and $\bm{A}_u^{(p)}$ \Big(resp.,~$\bm{A}_s^{(r)}$ and $\bm{A}_s^{(p)}$\Big) should yield a better prediction biasing $\Delta \mathcal{E}$ towards negative (resp.,~positive) values.  To yield a binary classification $\widehat{D} \in \{0, 1\}$, where $1$ denotes an unsafe response and $0$ denotes a safe response, we apply a decision threshold $\eta$:
\begin{equation} \label{decisioneq}
    \widehat{D} = 
    \begin{cases} 
    1, & \text{if } \Delta \mathcal{E} < \eta, \\ 
    0, & \text{if } \Delta \mathcal{E} \geq \eta. 
    \end{cases}
\end{equation}
The threshold $\eta$ serves as a hyperparameter to tune the balance between precision and recall, allowing for the optimization of the $F_1$ score across various LLM architectures.  The length of the response in \eqref{responselevel} can be adjusted as desired to classify individual sentences produced by the LLM or entire passages in response to a user prompt.

\section{Results and Discussions} \label{ressec}
We present our simulation results in three parts, covering the datasets used, the embedding models evaluated, and the experimental outcomes.
\subsection{Datasets}\label{sec:datasets}
To evaluate the performance of our DS approach on safety classification of LLM responses, we benchmark on the following three diverse datasets.
\paragraph{Aegis AI Content Safety Dataset 2.0 \cite{ghosh2025aegis}.}This dataset is designed to support the development of robust content safety guardrails for LLMs.  It comprises annotated human--LLM interactions drawn from diverse sources including Anthropic HH-RLHF, Do-Anything-Now (DAN) jailbreak prompts, and AI-assisted red-teaming datasets, with responses generated by Mistral-7B-v0.1. The dataset adheres to a comprehensive safety taxonomy of 12 categories (e.g., hate/identity hate, sexual, Violence, self-harm, criminal planning).  Safety labels are produced through a hybrid pipeline combining human annotations at the dialogue level with a multi-LLM jury system for response-level labels. In the original dataset paper \cite{ghosh2025aegis}, a Llama-3.1 Guard model was trained via parameter-efficient fine-tuning to achieve harmfulness $F_1$ scores of $80.8\%$.
\vspace{-0.38cm}
\paragraph{Synthetic CoT Safety Benchmark \cite{ai2-adapt-dev_synthetic-cot-safety}.} This dataset is designed to train and evaluate LLMs on their ability to refuse harmful requests through reasoned deliberation.  Each prompt/response pair has a potentially harmful user prompt with a structured model output that follows a Chain-of-Thought (CoT) with safety pattern: first a step-by-step safety reasoning trace that evaluates the harm potential of the request, followed by a clear refusal.\vspace{-0.38cm}
\paragraph{BeaverTails Dataset \cite{ji2023beavertails}.} This dataset is developed to support safety alignment research. It contains human prompts and LLM responses. Each sample belongs to one of 14 harm categories (including animal abuse, child abuse, discrimination, drug abuse, hate speech, privacy violation, self-harm, terrorism, and violence), together with an overall binary safety label.  A key feature of this dataset is that its prompts are human-written with responses generated by Alpaca-7B. This means that the prompts reflect the adversarial human intent rather than synthetic generation. This makes it particularly suitable for investigating whether our DS-based classification method can detect distinct embedding dynamics between human-written safe and unsafe prompts. A question-answering moderation model was trained in \cite{ji2023beavertails} then benchmarked to achieve a $F_1$ score up to $87.3\%$ on LLM response classification.

\subsection{Embedding Models}
We test our DS method using three top-performing embedding models from the HuggingFace leaderboard of the \href{https://huggingface.co/spaces/mteb/leaderboard}{Massive Text Embedding Benchmark} (MTEB) as described in Table \ref{tab:model_summary}. Working within limited computing resources, we selected models of varying sizes, from the lightweight $0.6$B-parameter \texttt{Qwen3-Embed} to the larger $8$B-parameter \texttt{Llama-3}, thereby balancing performance with practical hardware constraints. 
\begin{table}[h!]
\centering
\vspace{-0.1cm}
\caption{Benchmarked embedding models for DS safety detection.}
\label{tab:model_summary}

\begin{tabular}{@{}lllll@{}}
\toprule
\textbf{Model} & \textbf{Parameters} & \textbf{Dim} ($\bm{M}$)& \textbf{Max Context} & \textbf{Architecture Type} \\ \midrule
\texttt{Qwen3-Embed} \cite{li2026qwen3} & $0.6$\,B & $1024$ & $32768$ & Dense Decoder-Only \\
\texttt{Mistral} \cite{jiang2023mistral7b}& $7.2$\,B & $4096$ & $32768$ & Sparse Attention (SMoE) \\
\texttt{Llama-3} \cite{grattafiori2024llama}& $8.0$\,B & $4096$ & $8192$ & Dense Causal Decoder \\
\bottomrule
\end{tabular}

\end{table}

\subsection{Simulation Results}

We now present the classification results across all three datasets described in Section. \ref{sec:datasets}.  For each dataset, we report ROC curves for the three embedding models, followed by performance tables that compare response-only classification with joint prompt--response classification across varying sequence lengths $L \in\{1,50,100,150\}$.

\subsubsection{Aegis Dataset}

\begin{figure}[h!]
    \centering
    \subfloat[Qwen-Embed\label{fig:roc-aegis-qwen}]{{
    \includegraphics[scale=0.31]{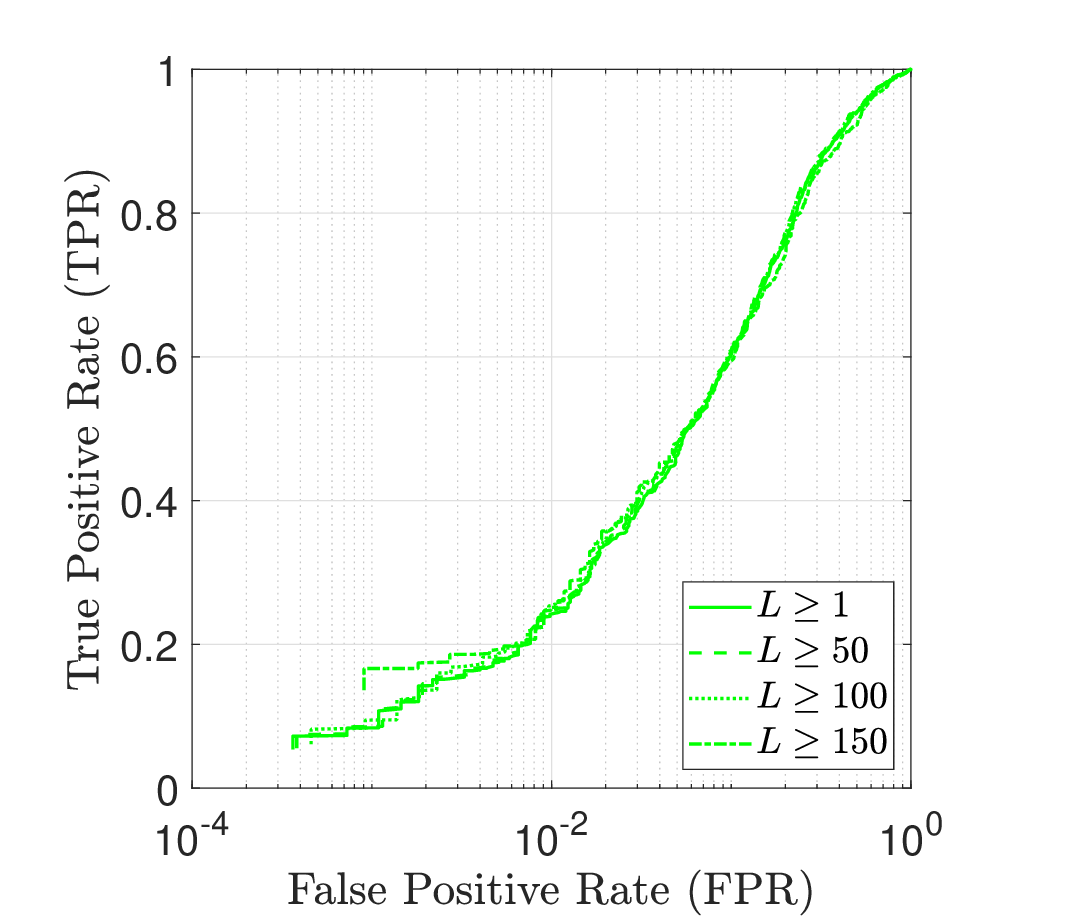} 
    }}
    \subfloat[Mistral\label{fig:roc-aegis-mistral}]{{
    \includegraphics[scale=0.31]{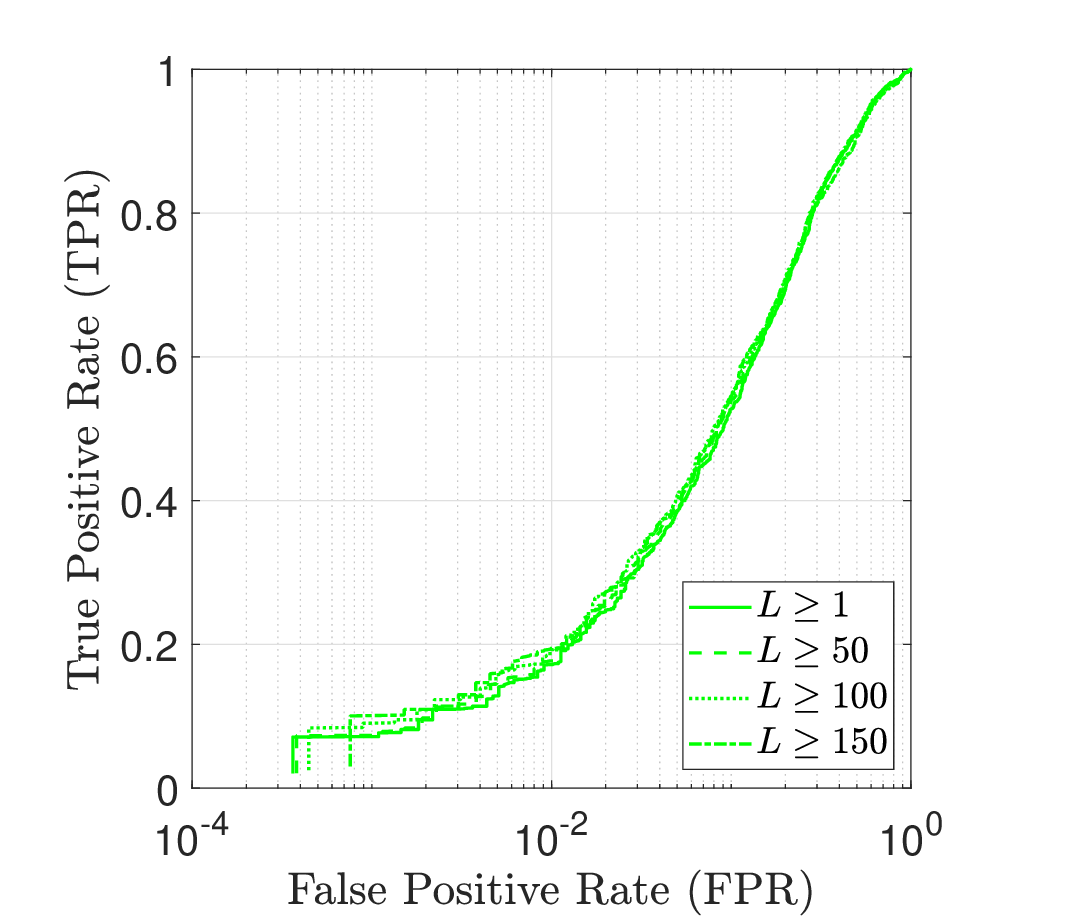} 
    }}
    \subfloat[Llama-3\label{fig:roc-aegis-llama3}]{{
    \includegraphics[scale=0.31]{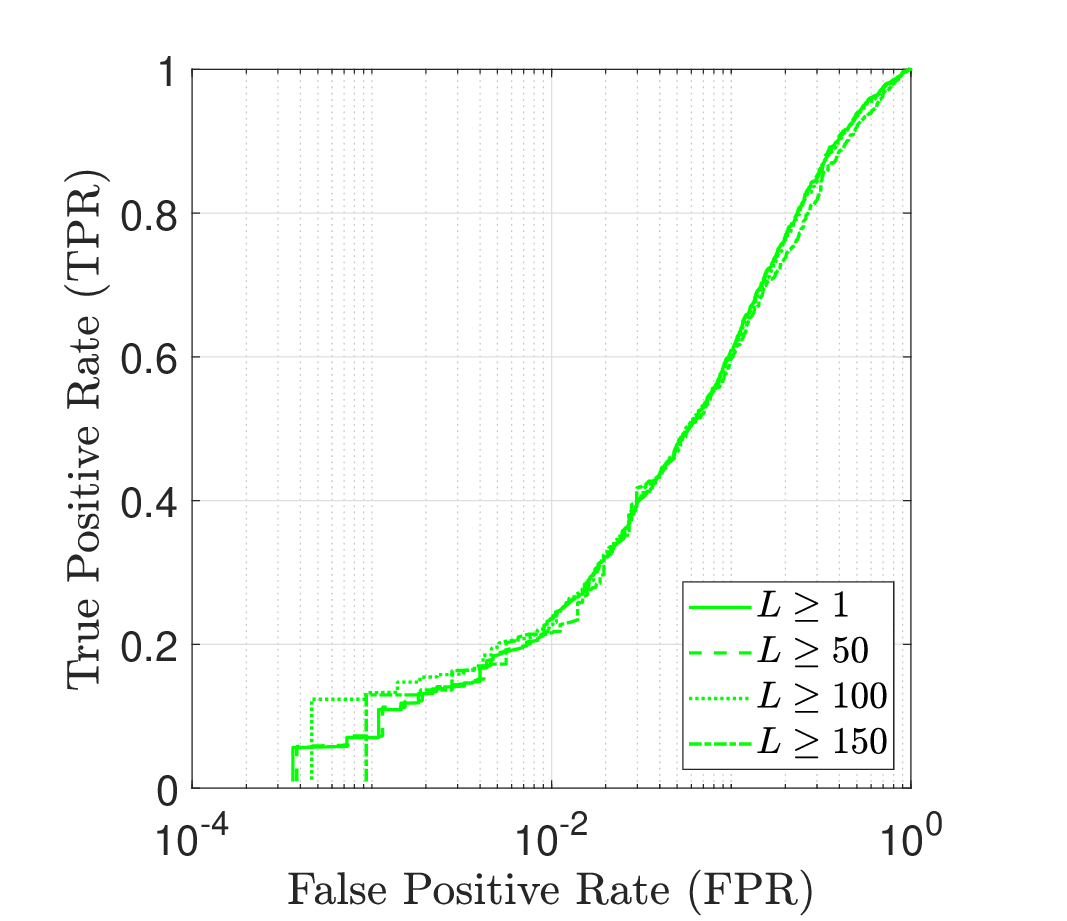} 
    }}
    \caption{ROC curves for different embedding models on the \texttt{Aegis} dataset.}
    \label{fig:Aegis_ROCs} 
\end{figure}

Figure~\ref{fig:Aegis_ROCs} shows the ROC curves for the three embedding models on the \texttt{Aegis} dataset.  All three models achieve clear separation from the diagonal, with \texttt{Llama-3} and \texttt{Qwen-Embed} exhibiting the strongest discriminative power.

Tables~\ref{tab:sequence-length-analysis-Aegis} and~\ref{tab:sequence-length-analysis-Aegis-prompt-4DMD} report classification performance on 12K test samples using response embeddings only and using separate DMDs for prompts and responses, respectively.  With response embeddings alone (Table~\ref{tab:sequence-length-analysis-Aegis}), \texttt{Qwen-Embed} and \texttt{Llama-3} achieve comparable $F_1$ scores around $72$--$73\%$, while \texttt{Mistral} trails at approximately $67$--$68\%$.  The addition of prompt embeddings (Table~\ref{tab:sequence-length-analysis-Aegis-prompt-4DMD}) reveals a significant shift: \texttt{Llama-3} emerges as the dominant model, achieving $F_1 = 77.0\%$ and accuracy of $80.2\%$ at $L \geq 150$, surpassing \texttt{Qwen-Embed} ($F_1 = 76.5\%$) and substantially outperforming \texttt{Mistral} ($F_1 = 68.8\%$).  This improvement is consistent across all sequence length thresholds, with \texttt{Llama-3} gaining $1.7$ percentage points in $F_1$ at $L \geq 1$ when prompt dynamics is incorporated, compared to $1.0$ points lost for \texttt{Qwen-Embed}.  These results are notable when compared to the baselines established in the Aegis~2.0 study: while the AegisGuard model~\cite{ghosh2025aegis} achieves $F_1 = 86.8\%$ on the Aegis test split through dedicated parameter-efficient fine-tuning on the training data, our DS method achieves $F_1 = 77.0\%$ ( and accuracy $80.5\%$) as a completely \emph{black-box} method that requires no task-specific training, i.e., by only the fitting of dynamical systems on a small set of embedding trajectories.

\begin{table}[h!]
    \centering
    %\vspace{-0.5cm}
    \caption{Classification performance on 12K test samples of the \texttt{Aegis} dataset over the number of tokens $L$ with \textit{responses' embeddings only}. The value of each cell denotes $F_1$ score/recall/accuracy.}
    \label{tab:sequence-length-analysis-Aegis}
    \begin{tabular}{l c c c c}
        \cmidrule{2-5}
        & \textbf{$L \geq 1$} & \textbf{$L \geq 50$} & \textbf{ $L \geq 100$} & \textbf{$L\geq 150$} \\
        \midrule
        \texttt{Qwen-Embed}  & \textbf{72.9}\;/\;\textbf{72.8}\;/\;72.3  &  \textbf{73.4}\;/\;\textbf{72.5}\;/\;\textbf{79.6}  &  \textbf{72.6}\;/\;\textbf{71.0}\;/\;\textbf{79.5}  &  72.4\;/\;69.1\;/\;78.5  \\
        \texttt{Mistral} & 67.5\;/\;63.8\;/\;76.5  & 68.1\;/\;63.9\;/\;76.8 & 67.7\;/\;62.9\;/\;77.0 &  67.7\;/\;61.0\;/\;77.0 \\
        \texttt{Llama-3} & 72.1\;/\;69.1\;/\;\textbf{79.6}  &  73.1\;/\;71.8\;/\;79.5  &  72.1\;/\;69.3\;/\;\textbf{79.5} &  \textbf{72.8}\;/\;\textbf{69.9}\;/\;\textbf{78.6} \\
        \bottomrule
    \end{tabular}
\end{table}

\begin{table}[h!]
    \centering
    \caption{Classification performance on 12K test samples of the \texttt{Aegis} dataset over the number of tokens $L$ with \textit{separate DMDs for prompts and responses' embeddings}. The value of each cell denotes $F_1$ score/recall/accuracy.}
    \label{tab:sequence-length-analysis-Aegis-prompt-4DMD}
    \begin{tabular}{l c c c c}
        \cmidrule{2-5}
        & \textbf{$L \geq 1$} & \textbf{$L \geq 50$} & \textbf{ $L \geq 100$} & \textbf{$L\geq 150$} \\
        \midrule
        \texttt{Qwen-Embed}  & 71.9\;/\;67.6\;/\;80.0  &  73.1\;/\;69.5\;/\;80.2  &  73.1\;/\;69.9\;/\;80.3  &  76.5\;/\;79.7\;/\;80.0  \\
        \texttt{Mistral} & 67.5\;/\;65.9\;/\;75.7  & 68.0\;/\;66.3\;/\;75.9 & 68.6\;/\;67.8\;/\;76.2 &  68.8\;/\;66.6\;/\;75.8 \\
        \texttt{Llama-3} & \textbf{73.8}\;/\;\textbf{72.7}\;/\;\textbf{80.3}  &  \textbf{74.4}\;/\;\textbf{73.2}\;/\;\textbf{80.5}  &  \textbf{74.2}\;/\;\textbf{73.3}\;/\;\textbf{80.4} &  \textbf{77.0}\;/\;\textbf{81.0}\;/\;\textbf{80.2} \\
        \bottomrule
    \end{tabular}
    \vspace{-0.2cm}
\end{table}

The dominance of \texttt{Llama-3} with prompt incorporation on Aegis can be explained as follows: the \texttt{Aegis} dataset is \emph{interaction-dependent}, meaning that safety violations arise from the interplay between user intent and model compliance.  \texttt{Llama-3}'s causal decoder architecture (through causal masking) captures the temporal dynamics of the prompt--response interaction more effectively, so that adding prompt embeddings significantly lowers the tracking error for safe samples while increasing the prediction error for unsafe samples.

\subsubsection{Synthetic CoT Safety Dataset}

\begin{figure}[h!]
    \centering
    \subfloat[Qwen-Embed\label{fig:roc-cot-qwen}]{{
    \includegraphics[scale=0.31]{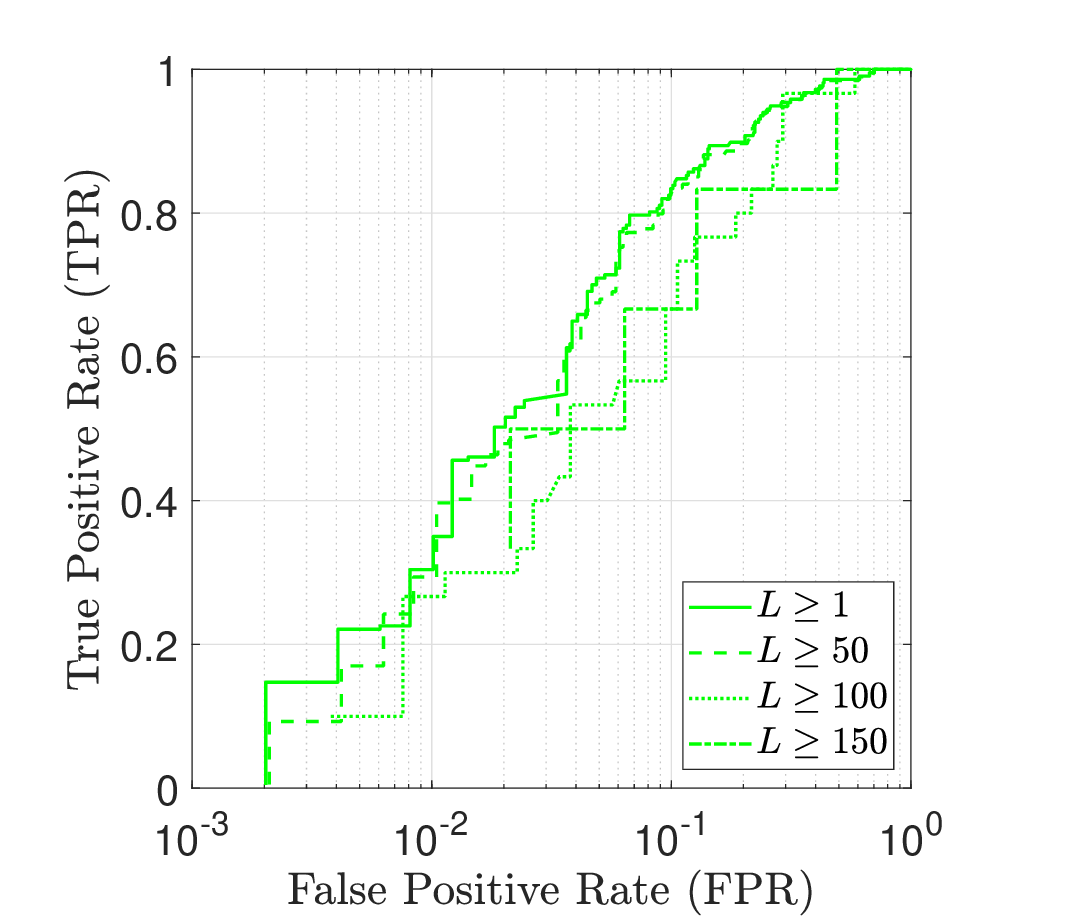} 
    }}
    \subfloat[Mistral\label{fig:roc-cot-mistral}]{{
    \includegraphics[scale=0.31]{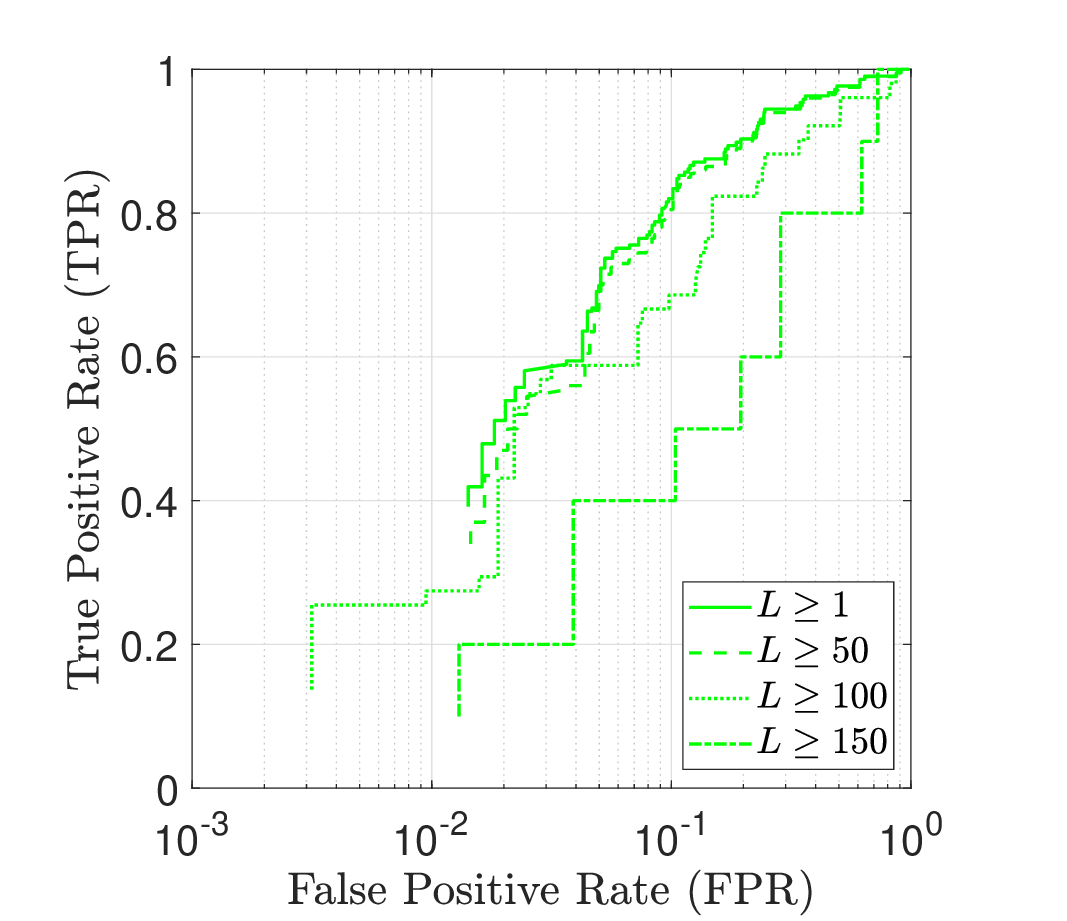} 
    }}
    \subfloat[Llama-3\label{fig:roc-cot-llama3}]{{
    \includegraphics[scale=0.31]{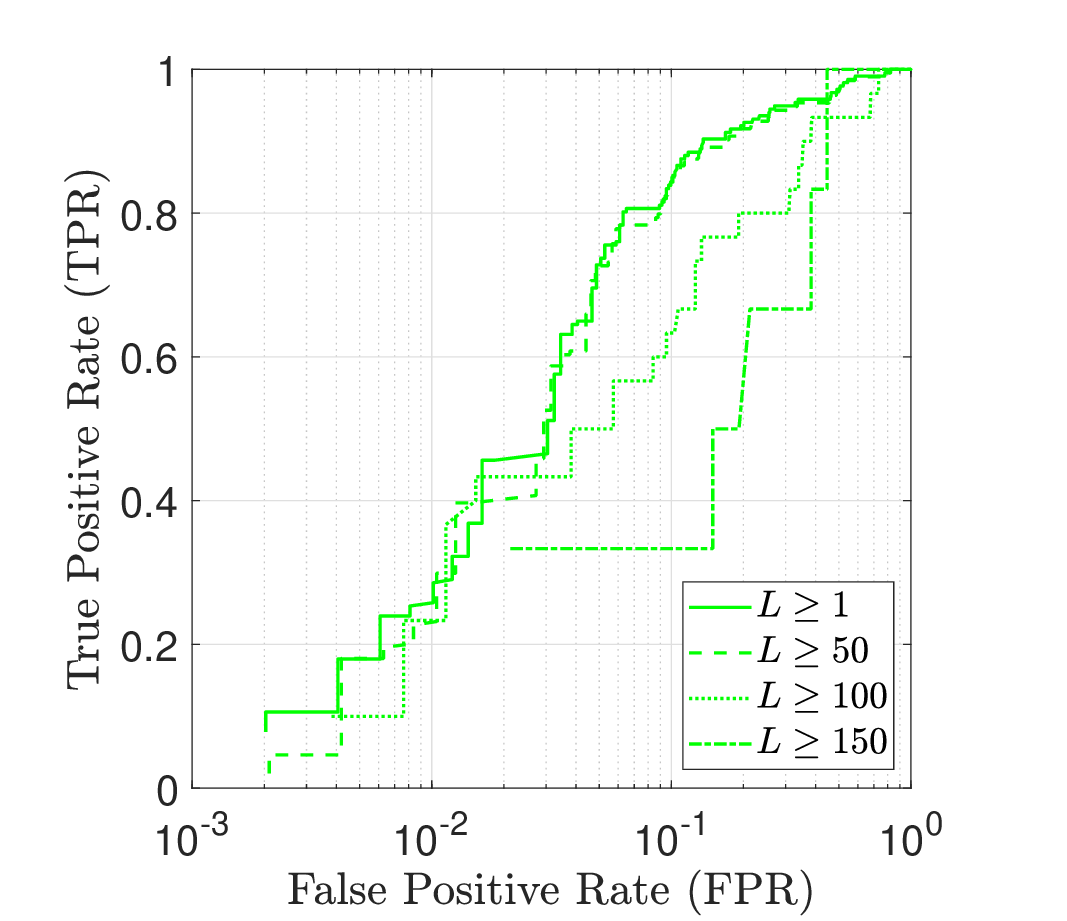} 
    }}
    \caption{ROC curves for different embedding models on the \texttt{Synthetic CoT Safety} dataset.}
    \label{fig:CoT_ROCs} 
\end{figure}

Figure~\ref{fig:CoT_ROCs} shows the ROC curves for the \texttt{Synthetic CoT Safety} dataset.  The overall discriminative power is high across all models at $L \geq 1$, while performance degrades more rapidly with increasing sequence length thresholds compared to the Aegis dataset. This reflects the smaller sample size at longer sequences.

Tables~\ref{tab:sequence-length-analysis-CoT} and~\ref{tab:sequence-length-analysis-CoT-prompt-4DMD} report performance on 710 test samples.  With response embeddings only, \texttt{Llama-3} leads at $L \geq 1$ with $F_1 = 83.0\%$ and accuracy $89.6\%$, followed closely by \texttt{Qwen-Embed} ($F_1 = 81.8\%$).  Adding prompt embeddings provides a consistent but modest boost: \texttt{Llama-3} reaches $F_1 = 83.7\%$ and accuracy $89.7\%$, while \texttt{Qwen-Embed} improves to $F_1 = 83.2\%$.  The performance drop at higher $L$ thresholds ($L \geq 100$ and $L \geq 150$) is attributable to the small number of test samples remaining at those lengths, which introduces high variance into the estimates.  Notably, the high accuracy across all settings (consistently above $88\%$) reflect the fact that this dataset has a pronounced class imbalance favoring safe samples, making accuracy alone a less informative metric than the $F_1$ metric.

\begin{table}[h!]
    \centering
    \caption{Classification performance on 710 test samples of the \texttt{Synthetic CoT Safety} dataset over the number of tokens $L$ with \textit{responses' embeddings only}. The value of each cell denotes $F_1$ score/recall/accuracy.}
    \label{tab:sequence-length-analysis-CoT}
    \begin{tabular}{l c c c c}
        \cmidrule{2-5}
        & \textbf{$L \geq 1$} & \textbf{$L \geq 50$} & \textbf{ $L \geq 100$} & \textbf{$L\geq 150$} \\
        \midrule
        \texttt{Qwen-Embed}  & 81.8\;/\;79.7\;/\;89.2  &  80.0\;/\;77.3\;/\;88.8  &  57.1\;/\;53.3\;/\;91.8  &  \textbf{60.0}\;/\;\textbf{50.0}\;/\;\textbf{92.5}  \\
        \texttt{Mistral} & 79.7\;/\;75.1\;/\;88.3  & 78.1\;/\;73.0\;/\;88.0 & \textbf{66.5}\;/\;\textbf{58.8}\;/\;91.6 &  47.1\;/\;40.0\;/\;90.0 \\
        \texttt{Llama-3} & \textbf{83.0}\;/\;\textbf{80.7}\;/\;\textbf{89.6}  &  \textbf{81.1}\;/\;\textbf{78.4}\;/\;\textbf{89.4}  &  55.3\;/\;43.3\;/\;\textbf{92.8} &  50.0\;/\;33.3\;/\;\textbf{92.5} \\
        \bottomrule
    \end{tabular}
\end{table}

\begin{table}[h!]
    \centering
    \caption{Classification performance on 710 test samples of the \texttt{Synthetic CoT Safety} dataset over the number of tokens $L$ with \textit{separate DMDs for prompts and responses' embeddings}. The value of each cell denotes $F_1$ score/recall/accuracy.}
    \label{tab:sequence-length-analysis-CoT-prompt-4DMD}
    \begin{tabular}{l c c c c}
        \cmidrule{2-5}
        & \textbf{$L \geq 1$} & \textbf{$L \geq 50$} & \textbf{ $L \geq 100$} & \textbf{$L\geq 150$} \\
        \midrule
        \texttt{Qwen-Embed}  & 83.2\;/\;85.3\;/\;89.4  &  81.8\;/\;83.5\;/\;89.3  &  59.6\;/\;46.7\;/\;\textbf{93.6}  &  \textbf{80.0}\;/\;\textbf{66.7}\;/\;\textbf{96.2}  \\
        \texttt{Mistral} & 81.1\;/\;77.9\;/\;88.9  & 79.6\;/\;76.0\;/\;88.6 & \textbf{63.0}\;/\;\textbf{56.9}\;/\;90.8 &  37.5\;/\;30.0\;/\;88.5 \\
        \texttt{Llama-3} & \textbf{83.7}\;/\;\textbf{86.2}\;/\;\textbf{89.7}  &  \textbf{82.4}\;/\;\textbf{84.5}\;/\;\textbf{89.6}  &  61.2\;/\;50.0\;/\;93.5 &  44.4\;/\;33.3\;/\;90.8 \\
        \bottomrule
    \end{tabular}
\end{table}

\subsubsection{BeaverTails Dataset}

\begin{figure}[h!]
    \centering
    \subfloat[Qwen-Embed\label{fig:roc-BeaverTails-qwen}]{{
    \includegraphics[scale=0.31]{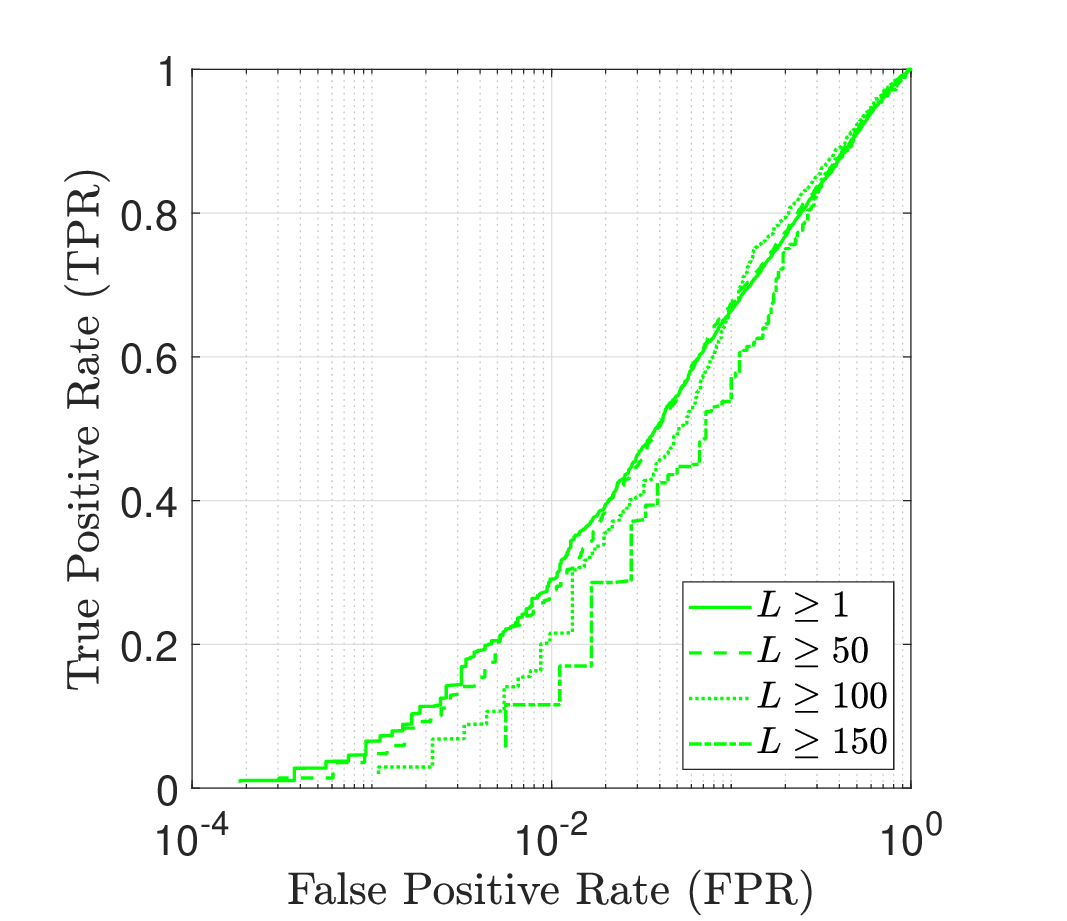} 
    }}
    \subfloat[Mistral\label{fig:roc-BeaverTails-mistral}]{{
    \includegraphics[scale=0.31]{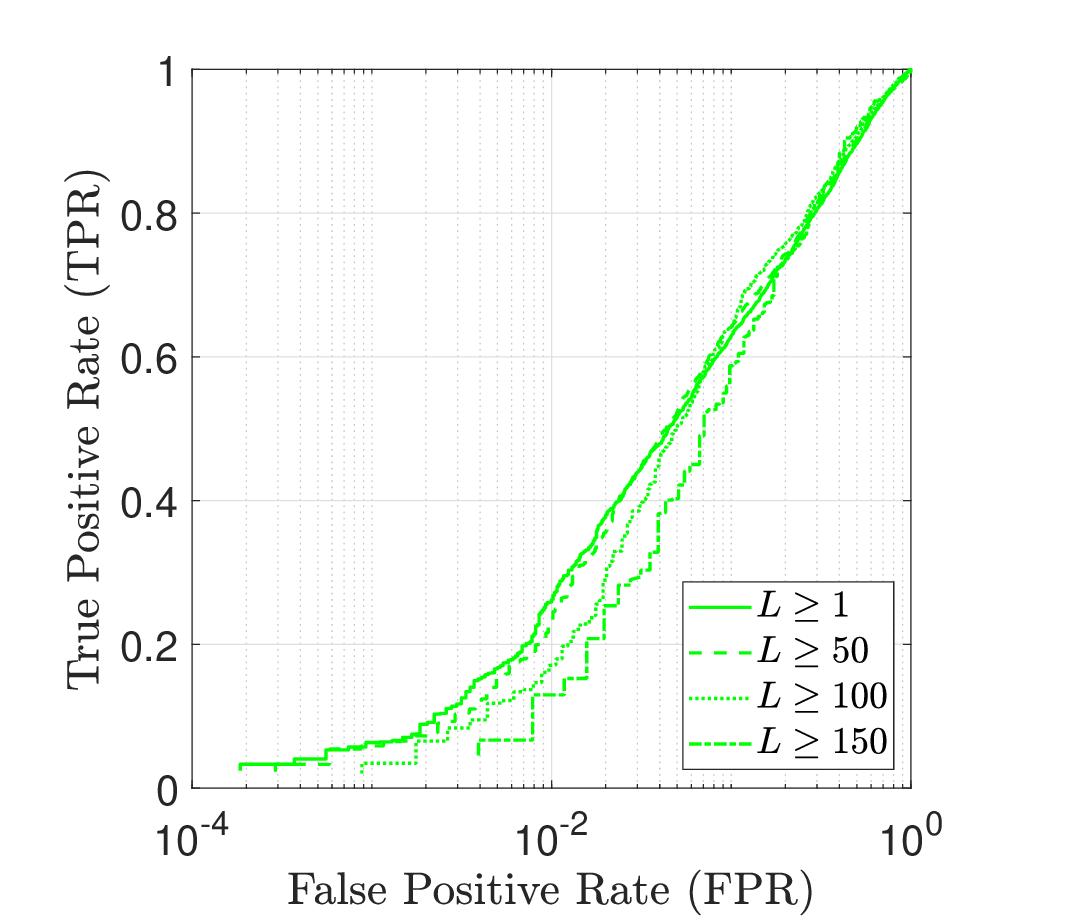} 
    }}
    \subfloat[Llama-3\label{fig:roc-BeaverTails-llama3}]{{
    \includegraphics[scale=0.31]{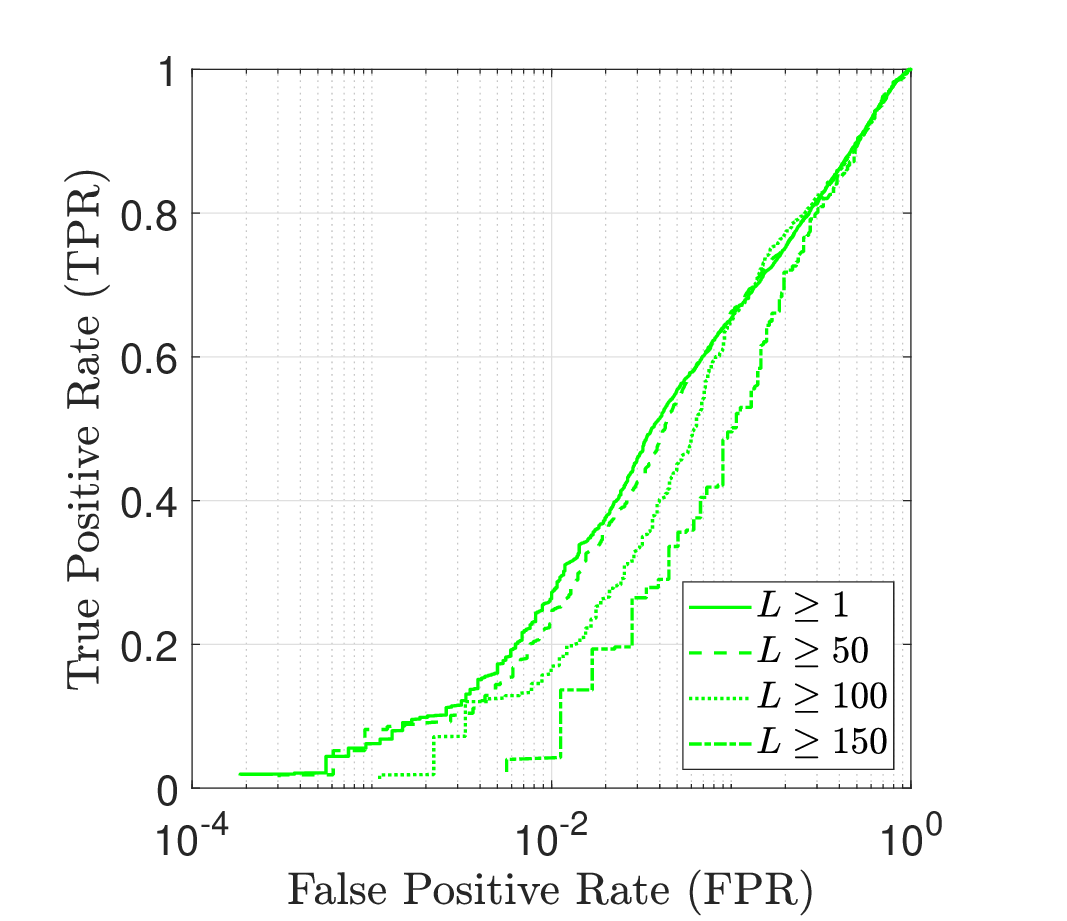} 
    }}
    \caption{ROC curves for different embedding models on the \texttt{BeaverTails} dataset.}
    \label{fig:BeaverTails_ROCs} 
\end{figure}

Figure~\ref{fig:BeaverTails_ROCs} shows the ROC curves for the \texttt{BeaverTails} dataset.  All three models achieve strong separation, with the curves shifting notably toward the upper-left corner as the sequence length increases. This is consistent with the hypothesis that longer token trajectories provide richer dynamical signatures for classification.

Tables~\ref{tab:sequence-length-analysis-BeaverTails},~\ref{tab:sequence-length-analysis-BeaverTails-prompt-4DMD}, and~\ref{tab:sequence-length-analysis-BeaverTails-prompt-4DMD-only} report performance on 12K test samples using response embeddings only, joint prompt--response embeddings, and prompt embeddings only, respectively.  With response embeddings only (Table~\ref{tab:sequence-length-analysis-BeaverTails}), \texttt{Qwen-Embed} is the strongest model, achieving $F_1 = 84.8\%$ at $L \geq 100$, while \texttt{Mistral} overtakes it at $L \geq 150$ ($F_1 = 85.6\%$).  Adding prompt embeddings (Table~\ref{tab:sequence-length-analysis-BeaverTails-prompt-4DMD}) provides consistent gains across all models, with \texttt{Qwen-Embed} reaching $F_1 = 86.7\%$ at $L \geq 150$---the highest $F_1$ score observed across all datasets and configurations in this study.

\begin{table}[h!]
    \centering
    \caption{Classification performance on 12K test samples of the \texttt{BeaverTails} dataset over the number of tokens $L$ with \textit{responses' embeddings only}. The value of each cell denotes $F_1$ score/recall/accuracy.}
    \label{tab:sequence-length-analysis-BeaverTails}
    \begin{tabular}{l c c c c}
        \cmidrule{2-5}
        & \textbf{$L \geq 1$} & \textbf{$L \geq 50$} & \textbf{ $L \geq 100$} & \textbf{$L\geq 150$} \\
        \midrule
        \texttt{Qwen-Embed}  & \textbf{80.0}\;/\;\textbf{78.2}\;/\;\textbf{78.4}  &  \textbf{81.8}\;/\;\textbf{81.6}\;/\;\textbf{78.7}  &  \textbf{84.8}\;/\;\textbf{82.9}\;/\;\textbf{80.3}  &  84.7\;/\;86.7\;/\;79.2  \\
        \texttt{Mistral} & 77.2\;/\;72.1\;/\;76.6  & 80.0\;/\;78.8\;/\;76.9 & 83.1\;/\;82.3\;/\;78.2 &  \textbf{85.6}\;/\;\textbf{90.5}\;/\;\textbf{79.5} \\
        \texttt{Llama-3} & 77.7\;/\;74.0\;/\;76.4  &  78.9\;/\;75.2\;/\;76.2  &  82.1\;/\;79.5\;/\;77.1 &  83.7\;/\;86.0\;/\;76.7 \\
        \bottomrule
    \end{tabular}
\end{table}

\begin{table}[h!]
    \centering
    \caption{Classification performance on 12K test samples of the \texttt{BeaverTails} dataset over the number of tokens $L$ with \textit{separate DMDs for prompts and responses' embeddings}. The value of each cell denotes $F_1$ score/recall/accuracy.}
    \label{tab:sequence-length-analysis-BeaverTails-prompt-4DMD}
    \begin{tabular}{l c c c c}
        \cmidrule{2-5}
        & \textbf{$L \geq 1$} & \textbf{$L \geq 50$} & \textbf{ $L \geq 100$} & \textbf{$L\geq 150$} \\
        \midrule
        \texttt{Qwen-Embed}  & \textbf{80.0}\;/\;\textbf{78.3}\;/\;78.4  &  \textbf{82.2}\;/\;\textbf{81.0}\;/\;\textbf{80.0}  &  \textbf{86.1}\;/\;85.4\;/\;\textbf{81.8}  &  \textbf{86.7}\;/\;\textbf{89.5}\;/\;\textbf{81.4}  \\
        \texttt{Mistral} & 78.2\;/\;74.1\;/\;77.6  & 81.0\;/\;78.5\;/\;78.1 & 85.0\;/\;86.0\;/\;80.1 &  86.3\;/\;86.6\;/\;81.5 \\
        \texttt{Llama-3} & 78.7\;/\;72.6\;/\;\textbf{78.5}  &  81.6\;/\;80.4\;/\;78.8  &  84.6\;/\;83.4\;/\;79.8 &  85.1\;/\;86.9\;/\;79.8 \\
        \bottomrule
    \end{tabular}
\end{table}

A distinctive feature of the \texttt{BeaverTails} dataset is that its prompts are human-authored, enabling us to investigate whether the embedding dynamics of human-written prompts carry independent safety information.  Table~\ref{tab:sequence-length-analysis-BeaverTails-prompt-4DMD-only} reports performance using \emph{prompt embeddings only}.  Even without any response information, prompt-only classification achieves $F_1 = 83.4\%$ at $L \geq 150$ with \texttt{Qwen-Embed}, demonstrating that human-authored safe and unsafe prompts do indeed yield dynamically distinguishable regions of the embedding space.  This finding suggests that the embedding dynamics of human language carry substantial safety-relevant information that can be exploited by our DS method.

\begin{table}[h!]
    \centering
    \caption{Classification performance on 12K test samples of the \texttt{BeaverTails} dataset over the number of tokens $L$ with \textit{separate DMDs for prompts' embeddings only}. The value of each cell denotes $F_1$ score/recall/accuracy.}
    \label{tab:sequence-length-analysis-BeaverTails-prompt-4DMD-only}
    \begin{tabular}{l c c c c}
        \cmidrule{2-5}
        & \textbf{$L \geq 1$} & \textbf{$L \geq 50$} & \textbf{ $L \geq 100$} & \textbf{$L\geq 150$} \\
        \midrule
        \texttt{Qwen-Embed}  & \textbf{73.1}\;/\;69.0\;/\;\textbf{71.0}  &  \textbf{77.9}\;/\;\textbf{76.7}\;/\;\textbf{73.9}  &  \textbf{82.4}\;/\;\textbf{85.2}\;/\;\textbf{75.9}  &  \textbf{83.4}\;/\;85.9\;/\;\textbf{77.0}  \\
        \texttt{Mistral} & 68.5\;/\;\textbf{74.4}\;/\;62.4  & 71.2\;/\;69.8\;/\;66.9 & 78.5\;/\;82.0\;/\;70.7 &  82.2\;/\;\textbf{92.4}\;/\;73.0 \\
        \texttt{Llama-3} & 66.6\;/\;61.2\;/\;66.2  &  71.1\;/\;65.4\;/\;68.8  &  80.0\;/\;82.6\;/\;72.6 &  79.4\;/\;80.1\;/\;72.4 \\
        \bottomrule
    \end{tabular}
\end{table}

When comparing the model rankings across the \texttt{Aegis} and \texttt{BeaverTails} datasets, adding prompt embeddings with \texttt{Aegis} causes \texttt{Llama-3} to overtake \texttt{Qwen-Embed} as the best-performing model (Table~\ref{tab:sequence-length-analysis-Aegis-prompt-4DMD}), whereas on \texttt{BeaverTails}, \texttt{Qwen-Embed} retains its advantage even after prompt embedding incorporation (Table~\ref{tab:sequence-length-analysis-BeaverTails-prompt-4DMD}).  This difference can be understood through the distinct nature of safety violations in each dataset.  On Aegis, violations are \textit{interaction-dependent}: the same prompt may be benign (e.g., an authorized security audit request) but the response can be unsafe if it provides actual exploit code rather than a refusal.  \texttt{Llama-3}'s causal decoder architecture captures this causal prompt--response interaction effectively, so that incorporating prompt dynamics significantly improves its tracking accuracy for safe samples.  On the \texttt{BeaverTails} dataset, violations are present at the LLM responses only: the unsafe response (e.g., graphic violence, hate speech, explicit drug instructions) is directly encoded in the embedding tokens, regardless of the prompt's phrasing.  In this case, \texttt{Qwen-Embed}'s semantic representations excel at mapping these safety-violating features into distinguishable embedding trajectories, and adding prompt embeddings provides only marginal additional information. It is worth noting that When classifying safety using prompt dynamics alone, the achieved classification performance is lower compared to tracking the LLM's response dynamics. This suggests that human-written inputs reveal significantly less to the Koopman operators about whether the subsequent response is safe or not.

\section{Conclusion} \label{concsec}

In this paper, we generalized a dynamical system method for LLM output classification to enforce LLM safety.  By fitting separate Koopman operators  for safe and unsafe embedding trajectories and classifying new LLM outputs through a differential residual score, our method achieves strong performance across three diverse safety benchmarks without requiring access to LLMs' internal variables, task-specific fine-tuning, or multiple stochastic samples.  The use of separate Koopman operators to track the token dynamics of the prompt and response led to consistent improvements compared to the case with response dynamics only, with large gains obtained on datasets where safety violations are interaction-dependent.

Our results reveal a nuanced relationship between embedding model architecture and the nature of safety violations.  Causal decoders like \texttt{Llama-3} excel when safety depends on the causal interaction between prompt and response, while compact semantic encoders like \texttt{Qwen-Embed} are better suited for detecting content-isolated violations where the unsafety is directly encoded in the response tokens.  This finding has practical implications for deploying DS-based safety monitors: the choice of embedding model should be informed by the expected distribution of safety violations in the target application.

Several directions for future work emerge from this study. Extending the binary safe/unsafe classification to multi-class prediction over specific harm categories (e.g., violence, hate speech, self-harm) would increase the practical utility of the method for content moderation systems that require actionable category labels. Moreover, exploring ensemble strategies that combine the complementary strengths of different embedding models may yield classifiers that are robust across diverse safety violation types.  Finally, a deeper theoretical investigation into the manifold structure of safe and unsafe embedding trajectories could provide formal guarantees on the separability conditions under which DS-based classification is expected to succeed.

\section*{Acknowledgment} 
This material is based upon the work supported by the National Science Foundation (NSF) under Grant No.~CMMI-2024111. This material is also based upon work co-supported by the U.S. Department of Energy, Office of Science, Office of Advanced Scientific Computing Research under Contract No. DE-AC05-00OR22725. This manuscript has been co-authored by UT-Battelle, LLC under Contract No. DE-AC05-00OR22725 with the U.S. Department of Energy. The United States Government retains and the publisher, by accepting the article for publication, acknowledges that the United States Government retains a non-exclusive, paid-up, irrevocable, world-wide license to publish or reproduce the published form of this manuscript, or allow others to do so, for United States Government purposes. The Department of Energy will provide public access to these results of federally sponsored research in accordance with the DOE Public Access Plan.

\FloatBarrier

\bibliography{myrefs}

\clearpage
\section*{Appendices}
\setcounter{section}{0}
\renewcommand{\thesection}{\Alph{section}}
\refstepcounter{section}
\subsection*{Appendix A: Koopman Operator Theory and Dynamic Mode Decomposition for Inferring Dynamics from Token Embeddings}\label{app:A}

Differential residual scores from Section \ref{residscore} are obtained by considering the token embedding dynamics of safe/unsafe prompts and responses. Following a strategy proposed in \cite{wilson2026lowcost},  we consider an LLM and its output as a dynamical system of the form
\begin{align} \label{dynsyst}
    \mathbf{x}_{k+1} &= F(\mathbf{x}_{k}), \nonumber \\
    \mathbf{q}_k &= G_1(\mathbf{x}_k), \nonumber \\
    \mathbf{y}_k &= H(\mathbf{x}_{k}) \equiv G_2(G_1(\mathbf{x}_k)),
\end{align}
where $\mathbf{x}_{k} \in \mathbb{R}^N$ are the state variables of an LLM immediately before the next token (i.e.,~activation maps), $F$ governs the state evolution, $\mathbf{q}_k$ is the token chosen by the LLM taken from its vocabulary, and $\mathbf{y}_k \in \mathbb{R}^M$ is the associated token embedding.  The nonlinear function $G_1$ maps the internal state to a token and $G_2$ maps the token to an embedding (e.g.,~via \texttt{Qwen}, \texttt{Mistral}, etc.).  Koopman operator theory \cite{mezi13,budi12} can be used consider the dynamics of observables, with the Koopman operator $K: \mathbb{R}^M \rightarrow \mathbb{R}^M$ defined according to \vspace{-0.1cm}
\begin{equation} \label{kaut}
    K H(\mathbf{x}_k) \equiv  H(F(\mathbf{x}_k)).
\end{equation}
 Despite the fact that the functions $F$ and $H$ are nonlinear, the Koopman operator is linear due to the linearity of the composition operator but generally infinite dimensional  \cite{budi12}, \cite{mezi13}. 

Dynamic mode decomposition (DMD) is a data-driven method that can be used to obtain a finite-dimensional approximation for the action of the Koopman operator \cite{kutz16}, \cite{schm10}, \cite{rowl09}, \cite{will15}.  Here, we consider the Extended DMD approach \cite{will15}.  To implement this strategy, we first lift the observables to a higher dimensional space 
 \begin{equation} \label{liftobs}
    \mathbf{z}_k = \begin{bmatrix} \mathbf{y}_k^\top  &  f_{\rm lift}^\top( \mathbf{y}_k)\end{bmatrix}^\top,
 \end{equation}
where $\mathbf{z}_k \in \mathbb{R}^{M+\gamma}$ is the lifted token embedding.  Here, $f_{\rm lift} \in \mathbb{R}^\gamma$ where $\gamma$ is the dimension of the lifting and $^\top$ indicates the vector transpose.  This initial lifting step generally yields a more accurate representation of the Koopman operator; common choices of lifted coordinates include polynomial combinations of observables and radial basis functions \cite{will15}.  In this work, we take the observable from \eqref{dynsyst} to be $\bm{y}_k = \bm{\Phi}^\top H(\mathbf{x}_k)$ where $\mathbf{\Phi} \in \mathbb{R}^{M \times 500}$ is comprised of the 500 most dominant SVD modes from the fitting data, as gauged by the singular values of the covariance matrix of the fitting data.  The lifting function $f_{\rm lift}$ takes polynomial combinations up to order 4 of a subset of the most dominant SVD modes.  

A set of snapshot pairs $s_k = (\mathbf{z}_k,\mathbf{z}_{k+1})$ is collected and arranged into matrices $\bm{X} = \begin{bmatrix} \mathbf{z}_1 & \dots & \mathbf{z}_{q}  \end{bmatrix}$ and $\bm{X}^+ = \begin{bmatrix} \mathbf{z}_2 & \dots & \mathbf{z}_{q+1}  \end{bmatrix}$ where $q$ is the number of snapshot pairs used. An approximation of the Koopman operator $\mathbf{z}_{k+1} = \mathbf{A} \mathbf{z}_k$ can be obtained according to
\begin{equation} \label{dmdfitting}
    \mathbf{A} = \mathbf{X}^+  \mathbf{X}^\dagger,
\end{equation}
where $^\dagger$ denotes the pseudoinverse.  Low-rank approximations of $\mathbf{A}$ can be obtained by truncating small magnitude SVD modes of $\mathbf{X}$ before taking the pseudoinverse.

\end{document}